\documentclass[letterpaper, 10 pt, conference]{ieeeconf}  

\IEEEoverridecommandlockouts                              

\usepackage{amsmath} 
\usepackage{amssymb}  
\usepackage{graphicx}
\usepackage{xcolor}

\title{\LARGE \bf
Learning Selective Communication for Multi-Agent Path Finding
}

\author{Ziyuan Ma$^{1*}$, Yudong Luo$^{2*}$ and Jia Pan$^3$

\thanks{$^{1}$School of Computing Science, Simon Fraser University, Canada
        {\tt\small ziyuan\_ma@sfu.ca}}%
\thanks{$^{2}$School of Computer Science, University of Waterloo, Canada
        {\tt\small yudong.luo@uwaterloo.ca}}%
\thanks{$^{3}$Department of Computer Science, The University of Hong Kong, Hong Kong, China
        {\tt\small jpan@cs.hku.hk}}%
\thanks{* indicates equal contribution.}%
}

\begin{document}

\maketitle
\thispagestyle{empty}
\pagestyle{empty}

\begin{abstract}

Learning communication via deep reinforcement learning (RL) or imitation learning (IL) has recently been shown to be an effective way to solve Multi-Agent Path Finding (MAPF). However, existing communication based MAPF solvers focus on broadcast communication, where an agent broadcasts its message to all other or predefined agents. It is not only impractical but also leads to redundant information that could even impair the multi-agent cooperation. A succinct communication scheme should learn \textit{which} information is relevant and influential to each agent’s decision making process. To address this problem, we consider a request-reply scenario and propose \textit{Decision Causal Communication} (DCC), a simple yet efficient model to enable agents to select neighbors to conduct communication during both training and execution. Specifically, a neighbor is determined as relevant and influential only when the presence of this neighbor causes the decision adjustment on the central agent. This judgment is learned only based on agent's local observation and thus suitable for decentralized execution to handle large scale problems. Empirical evaluation in obstacle-rich environment indicates the high success rate with low communication overhead of our method.

\end{abstract}

\section{Introduction}
Multi-Agent Path Finding is the problem of arranging a set of collision-free paths for a set of agents on a given graph. Although MAPF is NP-hard to solve optimally on graphs~\cite{yu2013structure} and even 2D grids~\cite{banfi2017intractability}, many optimal MAPF algorithms have been developed in recent years. Some reduce MAPF to other well-studied problems, e.g., ILP~\cite{yu2013planning} and SAT~\cite{surynek2016efficient}, others solve it with search-based algorithms, e.g., enhanced A* search~\cite{goldenberg2014enhanced,wagner2015subdimensional}, Conflict-Based Search~\cite{sharon2015conflict} and its improved variants~\cite{li2019improved}. However, the limitation of these centralized planning methods is that they do not scale well to a large number of agents.

RL and IL methods with decentralized execution have been applied to address this issue~\cite{sartoretti2019primal,wang2020mobile,riviere2020glas,liu2020mapper,li2020graph,li2021message,ma2021distributed}. During execution, each agent takes action based on its individual decision model with its own observation (may also include messages from other agents) as input, which avoids the scalability problem. For RL-based methods~\cite{sartoretti2019primal,wang2020mobile,liu2020mapper,ma2021distributed}, MAPF is generally modeled as a Markov Game~\cite{littman1994markov} with partial observability, where a reward function is designed for each agent, and the goal of each agent is to maximize its expected total return. Usually, expert guidance is applied to guide the RL during training in terms of behavior cloning~\cite{sartoretti2019primal}, shaped reward~\cite{liu2020mapper}, or heuristic map~\cite{ma2021distributed}. For IL-based methods~\cite{riviere2020glas,li2020graph,li2021message}, partial observability is still considered, but the objective is to minimize the divergence between expert demonstrations and generated trajectories, so that the learned policy can mimic the expert behavior. 

Recently, researchers focus on trainable communication channels between agents for MAPF, where extra information can be obtained during both training and execution to enhance multi-agent cooperation~\cite{li2020graph,li2021message,ma2021distributed}. However, these methods focus on broadcast communication in which messages will be transmitted to all other or predefined set of agents. For instance, in \cite{li2020graph} and \cite{li2021message}, the messages are broadcast to all other agents within a distance (communication radius). In \cite{ma2021distributed}, the central agent takes the messages from the two nearest neighbors after getting all the broadcast messages from its neighbors inside the field of view (FOV). Although empirical results have shown great improvements of communication based methods compared with non communication ones~\cite{ma2021distributed}, the drawback is that broadcast communication requires lots of bandwidth and incurs additional system overhead and latency in practice. More importantly, not every agent provides useful information for cooperation, and redundant information can even impair the learning process.

Plenty of methods on reducing communication overhead or learning selective communication have been proposed in the literature of RL~\cite{singh2018learning,kim2019learning,zhang2019efficient,zhang2020succinct,wang2020learning,ding2020learning}. But most of them are designed for centralized training framework on cooperative Markov Game, where agents share the same team reward. In contrast, each agent has its individual reward in MAPF, and thus more information needs to be swapped for cooperation. Therefore, selective communication is much more important in this case. To design a succinct communication protocol for MAPF, we consider a request-reply communication scenario where messages are updated in two stages, and propose \textit{Decision Causal Communication}, a simple yet efficient mechanism to enable agents to select other members in the swarm for communication during both training and execution. More specifically, before communication, each agent is capable to figure out which neighbors in its FOV are relevant and influential by just using local observation. The agent makes two temporary decisions based on its raw local observation and the modified local observation by masking out a neighbor. For any neighbor that can cause difference between these two temporary decisions, that neighbor is considered as relevant and influential, and the agent will send a request to that neighbor for communication purpose. This communication protocol is also naturally capable for decentralized execution. Empirical results show the high success rate with low communication overhead of our method compared with its counterparts.

\textbf{Contributions.} In this work, we propose DCC for MAPF. Unlike broadcast communication, DCC learns selective communication, and hence encourages agents to focus only on relevant information. We demonstrate that, on one hand, selective communication can remove temporal redundant messages, which is beneficial to multi-agent cooperation. On the other hand, communication overhead is greatly cut down due to the reduction of communication frequency.

\section{Related work}

\subsection{Reinforcement Learning based MAPF}\label{subsec:rl-mapf}
RL-based planners generally formulate MAPF as a multi-agent reinforcement learning (MARL) problem to learn decentralized policies for agents (robots). Compared with centralized planning methods, learning decentralized polices scales well to a large number of agents. Collision free policies are usually learned by guiding RL with constraints or demonstration data~\cite{everett2018motion,sartoretti2019primal,wang2020mobile,liu2020mapper}. Incorporating demonstration guidance with RL can be divided into two categories, using demonstrations generated by a centralized planner, or by a single-agent planner. For instance, a well known framework named PRIMAL~\cite{sartoretti2019primal} builds on the asynchronous advantage actor critic (A3C) network~\cite{mnih2016asynchronous} as its RL part and uses behaviour cloning to supervise the training of RL. So it requires demonstrations generated by a centralized planner named ODrM*~\cite{wagner2015subdimensional}. The limitation of using a centralized planner is that it requires solving a MAPF problem and is thus time-consuming, especially in a complex environment with a large number of agents. For methods using single-agent planner as guidance, MAPPER~\cite{liu2020mapper} and Globally Guided RL (G2RL)~\cite{wang2020mobile} use A* search for single-agent path generation and apply an off-route penalty if agents fail to follow the path. The potential issue is that, as single-agent shortest paths are usually not unique and not globally optimal for multi-agent environment, forcing agents to follow these paths by extra shaped rewards can mislead agents. 

To tackle the above issues, our previous method DHC~\cite{ma2021distributed} does not require a centralized planner. Although single-agent shortest paths are still adopted as guidance, DHC embeds all the potential choices of shortest paths as heuristic into the input of the model, instead of forcing agents to follow a specific path. Thus, no special shaped rewards are required. 

Even though empirical results have shown that enhancing RL with guidance can help to learn collision free policies for MAPF, the cooperation among agents is not directly modeled. Communication via graph convolution is a promising way to achieve multi-agent cooperation, where an agent aggregates information from its neighbors, including itself, and makes decisions based on this augmented information~\cite{das2019tarmac,jiang2020graph,niu2021multi}. This idea is recently deployed for MAPF. By treating each agent in the environment as a node and connecting neighboring nodes (agents) if they are inside the FOV of each other, a graph is formulated by DHC and multi-head dot-product attention~\cite{vaswani2017attention} serves as the convolutional kernel to compute interactions among agents. Similar ideas are applied by the work described in~\cite{li2020graph} and Message-Aware Graph Attention neTwork (MAGAT)~\cite{li2021message}, where the communication part is a graph neural network (GNN)~\cite{scarselli2008graph}. However, it is worth noting that both \cite{li2020graph} and \cite{li2021message} are IL-based methods, which train the model to imitate the demonstrations generated by Conflict-Based Search (CBS)~\cite{sharon2015conflict} and Enhanced CBS~\cite{barer2014suboptimal}, respectively. Thus, MAPF instances still need to be pre-solved while extreme scenarios may fail to be solved and collected into the training data. Also, none of these communication based methods consider the efficiency of communication, resulting in high communication overhead and latency.

\subsection{Efficient and Selective Communication for MARL}
Most existing works on communication in MARL, including those described above, focus on broadcast communication, i.e., broadcasting the messages to all other or predefined agents. To improve efficiency, a number of strategies have been proposed. Individualized Controlled Continuous Communication Model (IC3Net)~\cite{singh2018learning} considers that full communication is not always necessary and agents can determine whether to send messages to others indicated by a gating function. However, this gating function can block all communication channels of an agent. Cases where an agent's message is instructive to some agents but useless or harmful to others can not be handled. Schedule Communication (SchedNet)~\cite{kim2019learning} only allows a limited number of agents to broadcast messages, who are chosen by some importance weights assigned. The shortcoming is that a central scheduler is required to gather all individual weights and decide which agents should be entitled to broadcast their messages. Variance Based Control (VBC)~\cite{zhang2019efficient} lets agents transmit their messages only when the variance of the message vector is high (high variance implies that the message is informative). And agents decide whether to request messages based on local ambiguity. Temporal Message Control (TMC)~\cite{zhang2020succinct} stores agent messages in the buffer and new messages are sent out only when they contain relatively new information compared with old ones. Nearly Decomposable Q-functions (NDQ)~\cite{wang2020learning} optimizes communication via minimizing the entropy of messages between agents. Individually Inferred Communication (I2C)~\cite{ding2020learning} learns a prior network, which takes local observation and identity of another agent as input, to predict a belief of whether to communicate. All VBC, TMC, NDQ and I2C are designed on the centralized training methods that factorize the joint action-value function, such as QMIX~\cite{rashid2018qmix}, different from MAPF settings.

\section{Problem formulation}\label{sec:env}

\subsection{MAPF Problem Definition}
MAPF has many variants as summarized in~\cite{stern2019multi}. In this paper, we consider the classical MAPF case defined in~\cite{stern2019multi} that (a) each agent performs an action in every time step, and may cause vertex and swapping conflicts, (b) agents "stay at target" until all agents have reached their goals, and (c) optimizes the sum of cost. This definition differs from the environments used in PRIMAL and MAPPER, which are much simpler. In PRIMAL, agents take actions in turn (only one agent takes action at a time), although we can treat this as all agents act simultaneously but only one agent can move while all others choose no-op action. In MAPPER, agents are removed from the environment upon reaching their goal locations, which simplifies the problem, as there will be no livelocks when an agent reaches its goal but obstructs other agents from getting to their goals. We also notice that the environment in \cite{damani2021primal} is specially designed where obstacles compose narrow corridors that only allow one agent to pass at a time. We do not consider an environment like this.

Formally, we define MAPF as follows. Given an undirected graph $G=(V,E)$ and a set of $n$ agents, each agent indexed by $i$ has a unique start vertex $s_i\in V$ and a unique goal vertex $g_i\in V$. Time is discretized into time steps. At each discrete time step, each agent can either \textit{move} to an adjacent vertex or \textit{wait} at its current vertex. A path for $i$-th agent is a sequence of adjacent (indicating a moving) or identical (indicating a waiting) vertices starting at $s_i$ and terminating at $g_i$. Agents remain at their goal vertices ($g_i$) after they complete their paths. There are two kinds of collisions, either vertex collision or edge collision. A vertex collision is a tuple $\langle i,j,v,t \rangle$ where $i$-th and $j$-th agents reaching at the same vertex $v$ at time $t$. An edge collision is a tuple $\langle i, j, u, v, t\rangle$ where $i$-th and $j$-th agents traverse the same edge $(u,v)$ in opposite directions at time $t$. A solution to MAPF is a set of collision-free paths, one for each agent. The quality of a solution is measured by the sum of arrival time of all agents at their goal vertices. 

\begin{figure}[tpb]
\centering
\includegraphics[scale=0.19]{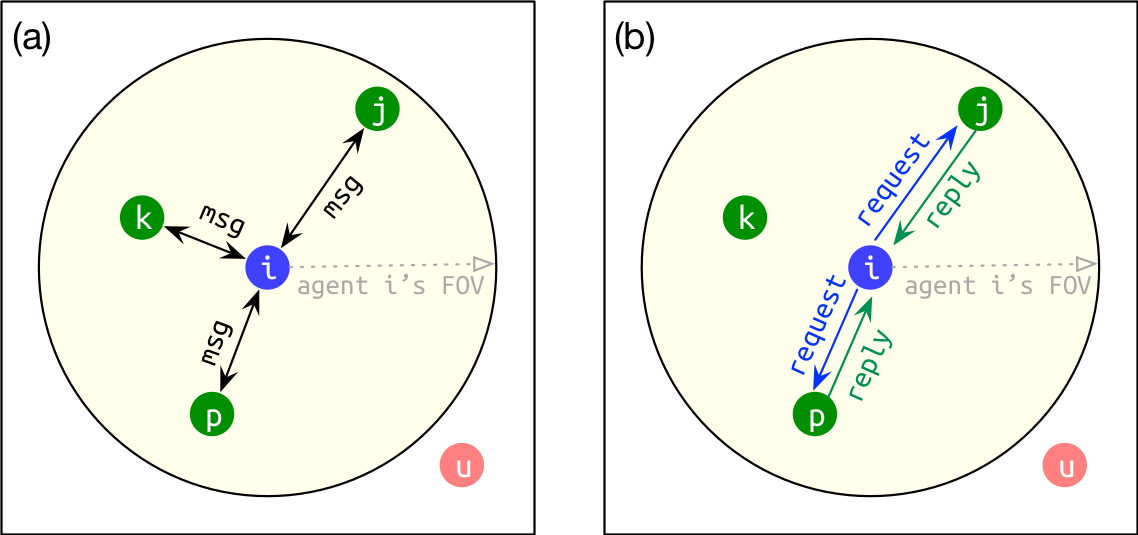}
\caption{(a) Broadcast communication: agent $i$ broadcasts its message to all neighboring agents $j$, $k$, and $p$. And agents $j$, $k$, $p$ will also broadcast their messages to agent $i$. If predefined scope is set, agent $i$ can decide whose message to be used in calculation, e.g., DHC takes the nearest two agents. There is no communication between agent $i$ and $u$, as agent $u$ is outside agent $i$'s FOV. (b) Request-reply based selective communication: agent $i$ omits irrelevant agent $k$, and only communicates with relevant agent $j$ and $p$. After determining which agent to communicate, the communication is conducted in two rounds. At the first round, agent $i$ sends out messages (requests) to agent $j$ and $p$. At the second round, agent $j$ and $p$ reply with their updated messages to agent $i$. }
\label{fig:req-rep}
\vspace{-0.17in}
\end{figure}

\subsection{Environment Setup}
In line with standard MAPF tasks, we focus on 2D 4-neighbor grids where agents, goals, and obstacles occupy one grid cell respectively. Formally, the map is a $m\times m$ matrix, where $0$ represents a free location and $1$ is an obstacle. Each map is chosen $n$ start positions and $n$ corresponding goal positions for $n$ agents from free locations. We make sure each goal is reachable from its start point and there is no overlap among $2n$ selected positions.  At each time step, agents move simultaneously to neighboring locations or wait at their current locations. Thus, the size of the action space is 5 (move to four directions or stay still). Agents may hit obstacles or conflict with others during simulation. We handle the collision by recursively resetting the related agents to previous states until no collision exits. 

We consider a partially observable environment, which is more compatible with the real world problem. Each agent can only access the information inside its FOV with size $\ell\times\ell$ ($\ell<m$), where $\ell$ is an odd number to make sure agents are at the center of FOV.

The reward design is adopted from DHC~\cite{ma2021distributed}, shown in Table~\ref{table:reward}. Generally, negative rewards are given to agents for each movement to facilitate goal reaching. No shaped reward is required in our method.

\section{Selective communication for MAPF}

In this paper, we propose DCC \footnote{Code available at https://github.com/ZiyuanMa/DCC}, a simple yet efficient mechanism to reduce communication overhead in multi-agent systems. DCC can be instantiated by independent Q-learning~\cite{tan1993multi}, or any framework of centralized training and decentralized execution, such as QMIX~\cite{rashid2018qmix} and VDN~\cite{sunehag2018value}. We consider a request-reply scenario different from the traditional setup where each agent only sends out an almost information-less indicator, e.g. a scalar value, as the request signal. Instead, in our setup, each agent sends out a request signal with rich information including its own messages along with the relative positions of neighbors. In this way, after receiving the request, the agents being requested can immediately benefit from this query by collecting some information from the query agent.  Fig.~\ref{fig:req-rep} shows the difference between broadcast and our request-reply based selective communication. In the following part, We first introduce DCC communication mechanism and then instantiate DCC with independent Q-learning.


\begin{table}[tbp]
\caption{Reward Function Design}
\label{table:reward}
\begin{center}
\resizebox{0.65\columnwidth}{!}{
\begin{tabular}{|c|c|}
\hline
\textbf{Actions} & \textbf{Reward}\\
\hline
Move (Up/Down/Left/Right)      & -0.075  \\
\hline
Stay (on goal, away goal) & 0, -0.075 \\
\hline
Collision (obstacle/agents) & -0.5 \\
\hline
Finish            & 3 \\
\hline
\end{tabular}
}
\end{center}
\vspace{-0.17in}
\end{table}

\begin{figure*}
\centering
\includegraphics[scale=0.26]{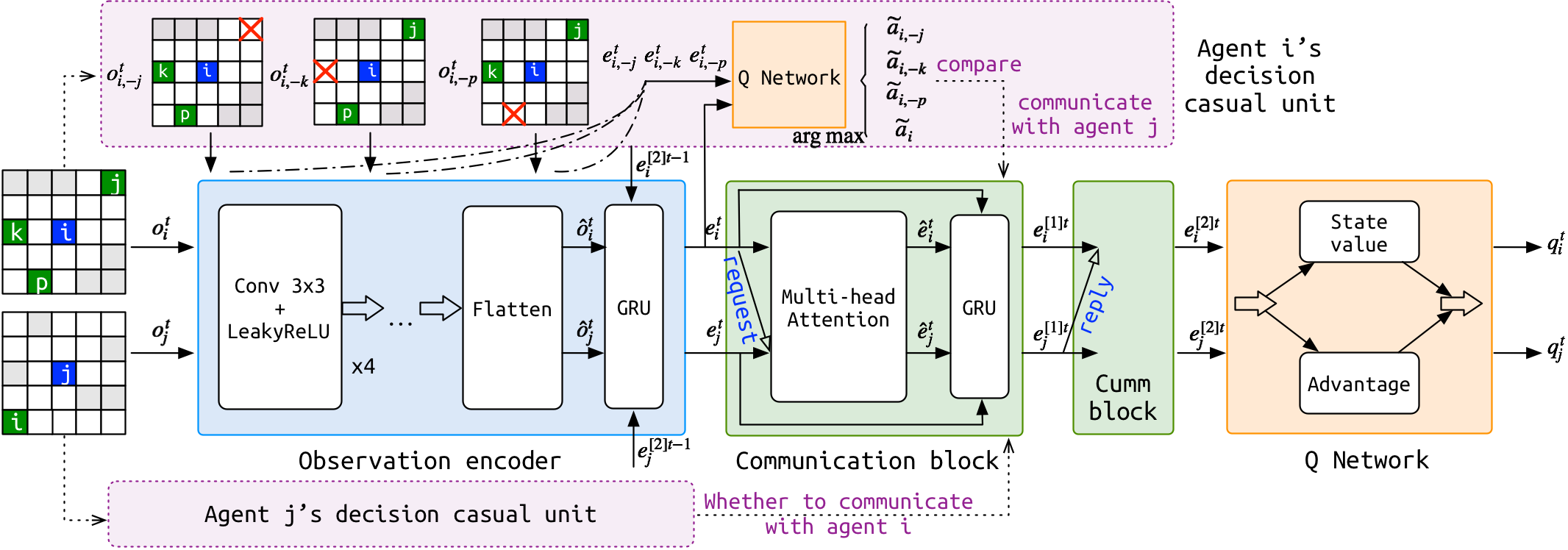}
\caption{System flow of DCC. It contains four modules: observation encoder (blue), decision casual unit (purple), communication block (green), and Q network (orange). An example of agent $i$'s FOV is shown by a $5\times 5$ grid, where the blue color represents the agent itself, the green color represents neighbors, and the gray color represents obstacles. Agent $i$'s decision casual unit constructs three modified local observations by masking each neighbor inside the FOV, represented by a red cross. The embeddings by observation encoder for the raw and modified local observations are fed to Q network directly to get temporary decisions, which are further used to decide which neighbor is relevant for communication.}
\label{fig:model}
\vspace{-0.1in}
\end{figure*}

\subsection{Decision Causal Communication}\label{subsec:DCC}
Our communication mechanism is motivated by I2C~\cite{ding2020learning}, where causal inference is utilized to derive a threshold to trigger communication. Specifically, I2C defines the causal effect of agent $j$ to agent $i$ as the KL-divergence~\cite{kullback1997information} between two decision probabilities, $D_{KL}( P(a_i|\boldsymbol{a}_{-i}, \boldsymbol{o}) || P(a_i|\boldsymbol{a}_{-ij}, \boldsymbol{o}) )$, where $\boldsymbol{o}$ is the joint observations, $\boldsymbol{a}_{-i}$ denotes the joint actions except for agent $i$, and similar meaning for $\boldsymbol{a}_{-ij}$. The magnitude of this divergence indicates how much agent $i$ will adjust its policy if taken into account agent $j$'s policy. If the divergence is small, agent $i$ chooses not to communicate with $j$ as that will hardly affect its own policy.

However, I2C is only effective for centralized training algorithms as it requires to know the joint observations and actions of all other agents. In real world problems such as MAPF, usually an agent with partial observability can only observe the existence of other agents instead of their policies. Although agents can broadcast their actions, that will definitely increase the communication overhead. In practice, if the policy of an agent requesting communication conditions on the action of the communicated agent, the circular dependencies can occur. At last, setting a reasonable threshold for that KL-divergence highly depends on empirical experiments and may vary in different problems.

Thus, to design an appropriate communication protocol for MAPF, in this work, we restrict the strategy in I2C by triggering communication between agent $i$ and $j$ only when the \textit{existence} of agent $j$ causes the policy adjustment on agent $i$, and we call this \textit{Decision Causal Communication}. To get an intuition of the proposed protocol, consider the learning based MAPF methods discussed in Section~\ref{subsec:rl-mapf}, where most of them guide the RL agent with global or local optimal demonstrations. In order to diminish the communication cost, an agent should follow the demonstration guided policy as often as possible without communicating with others, if the existence of other agents in its FOV will not affect its current policy. Otherwise, the agent communicates with those neighbors who will individually affect its policy. Although the policy may be adjusted after communication and we can not guarantee the adjusted policy is always better than unchanged policy during simulation, training the communication part along with RL in an end to end manner to maximize the expected total return will force the communication to be helpful and performed only when it is necessary.

Formally, this communication protocol can be formulated as follows. Suppose agent $j$ can be seen by agent $i$ (agent $j$ is inside the FOV of agent $i$), and the relative position of agent $j$ in $i$'s FOV is $(x_j, y_j)$. Agent $i$ gets its local observation $o_i$ from the environment and constructs another modified observation $o_{i,-j}$ ($-j$ means without $j$) by setting the information at agent $j$'s position by some special value, such as zero: $o_{i,-j}\leftarrow o_{i}(x_j, y_j)=0$. Then agent $i$ uses its local decision model, without communication, to predict two temporary actions $\widetilde{a}_i$ and $\widetilde{a}_{i,-j}$ based on $o_i$ and $o_{i,-j}$, respectively. If these two actions match with each other, implying that the existence of agent $j$ will not affect $i$'s policy, agent $i$ will not request message from agent $j$. Otherwise, agent $i$ sends a request to agent $j$ via a communication channel, and agent $j$ will respond with a message. In this work, we consider messages are updated in both request and reply stages, which is discussed in the following subsection. The final aggregated messages are further used by policy or Q-network to generate the final actions.

\subsection{Deep Q-Learning Model Design}
We instantiate DCC with independent Q-learning, namely the model is designed and trained from a single agent's perspective by treating other agents as part of the environment. We borrow some architecture design from DHC. The whole model of DCC consists of four main components, including observation encoder, decision causal unit, communication block, and Q-network. The model architecture is shown in Fig.~\ref{fig:model}. We begin from the model input and describe the four components one by one.

\paragraph{Local observation input} At time step $t$, agent $i$ gets a matrix with the shape $\ell\times \ell\times 6$ as input, denoted by $o_i^t$. The first channel is a binary matrix representing the obstacles inside the FOV. The second channel is a binary matrix indicating the locations of other agents if within the FOV. The other four are heuristic channels proposed by DHC~\cite{ma2021distributed}, to encode multiple choices of single-agent shortest path. We refer readers to the DHC paper for further details.

\paragraph{Observation encoder} The encoder contains \textit{four} stacked convolutional layers with a GRU~\cite{chung2014empirical}. The local observation $o^t_i$ is first encoded into $\hat{o}_i^t$ by four convolutional layers. Then by adopting the last step communication outcome $e^{[2]t-1}_i$ as the hidden state, the GRU gets $\hat{o}_i^t$ and generates the intermediate message $e^t_i$.

\paragraph{Decision causal unit} This unit depends on the observation encoder to embed modified observations, and the Q-network (will be discussed later), to get temporary actions. Let $\mathbb{B}_i$ denote all neighbors of agent $i$. Given $o^t_i$, $o^t_{i,-j}$ is constructed according to Section~\ref{subsec:DCC} for all $j\in\mathbb{B}_i$. Each $o^t_{i, -j}$ is passed through the observation encoder as discussed above to get an embedding $e^t_{i,-j}$. By skipping the communication, $e^t_i$ and $\{e^t_{i,-j}\}_{j\in\mathbb{B}_i}$ are directly fed into the Q-network to compute Q-values. Actions $\widetilde{a}^t_i$ and $\{\widetilde{a}^t_{i,-j}\}_{j\in\mathbb{B}_i}$ are inferred by applying argmax over Q-values. Based on these actions, the communication scope of agent $i$ is
\begin{align}
\mathbb{C}_i = \{j|\widetilde{a}^t_i\neq \widetilde{a}^t_{i, -j}\}_{j\in\mathbb{B}_i}.
\end{align}
Note that temporary actions $\widetilde{a}^t_i$ and $\{\widetilde{a}^t_{i,-j}\}_{j\in\mathbb{B}_i}$ are only used to decide the communication scope, not the final action to be executed.

\paragraph{Communication block} This block is the graph convolution with multi-head dot-production as the convolutional kernel~\cite{vaswani2017attention} followed by a GRU. We regard the request-reply as a two-round communication. After defining the communication scope $\mathbb{C}_i$, at the first round, agent $i$ sends request information to all agent $j\in\mathbb{C}_i$. This information includes its own message $e^t_i$ as well as the relative position of $i$'s neighbors inside $i$'s FOV, denoted as $l^t_i$. The relative position of each neighbor is originally represented as a one-hot vector, i.e. the vector length is  $\ell\times\ell$, and further embedded by a neural network layer. At the first round, agent $j$ may receive many requests from different agents, and we define this request receiving scope for agent $j$ as
\begin{align}
\bar{\mathbb{C}}_j = \{i|j\in \mathbb{C}_i\}. 
\end{align}

Then each agent $j$ who receives requests will update its message using multi-head attention. Let $\bar{\mathbb{C}}_{j+}$ denote the set $\{j,\bar{\mathbb{C}}_j\}$. For every agent $\bar{i}\in\bar{\mathbb{C}}_{j+}$, agent $\bar{i}$'s own message $e^t_{\bar{i}}$ is projected to \textbf{Q}uery by matrix $\boldsymbol{W}^h_Q$, while the concatenation of $e_{\bar{i}}^t$ and $l_{\bar{i}}^t$ is projected to \textbf{K}ey and \textbf{V}alue by matrix $\boldsymbol{W}^h_K$ and $\boldsymbol{W}^h_V$ in each independent attention head $h$. The relation between agent $j$ and $\bar{i}\in\bar{\mathbb{C}}_{j+}$ in $h$-th attention head is computed as 
\begin{align}
\mu^h_{j\bar{i}}=\mathrm{softmax}\left(\frac{\boldsymbol{W}^h_Q e^t_j\cdot (\boldsymbol{W}^h_K [e^t_{\bar{i}},l^t_{\bar{i}}])^{\top} }{\sqrt{d_K}}\right),
\end{align}
where $d_K$ is the dimension of \textbf{K}eys. The \textbf{V}alues are weighted summed by weights $\mu^h_{j\bar{i}}$ over $\bar{\mathbb{C}}_{j+}$ at each head $h$. And all head outputs are concatenated over $\mathcal{H}$ heads to pass through a neural network layer $f_o$ for the ﬁnal output
\begin{align}
\hat{e}^t_j=f_o\left(\mathrm{concat}\left(\sum_{\bar{i}\in\bar{\mathbb{C}}_{j+}} \mu^h_{j\bar{i}}\boldsymbol{W}^h_V[e^t_{\bar{i}},l^t_{\bar{i}}], \forall h\in\mathcal{H}\right)\right). 
\end{align}
The message $\hat{e}^t_j$ and the initial message $e^t_j$ are first aggregated by a GRU to generate the first round outcome $e^{[1]t}_{j}$. Then $e^{[1]t}_{j}$ acts as the initial message for the next round.

At the second round, agents who received requests will reply with their updated messages along with the relative position of their neighbors, i.e. agent $i$ will receive $e^{[1]t}_j$ and $l_j^t$ from all agent $j\in\mathbb{C}_i$. Let $\mathbb{C}_{i+}$ denote the set $\{i,\mathbb{C}_i\}$. Agent $i$ will update its message in the same manner as defined in Equations 3-4 by replacing $j$ with $i$, and replacing  $\bar{i}\in\bar{\mathbb{C}}_{j+}$ with $\bar{j}\in\mathbb{C}_{i+}$. Denote the message generated by Equations 3-4 at this round as $\hat{e}^{[1]t}_i$. Finally, $\hat{e}^{[1]t}_i$ and $e^{[1]t}_j$ are aggregated by a GRU to output $e^{[2]t}_i$, the final outcome message of the second round.

\paragraph{Q-network} The message $e^{
[2]t}_i$ is adopted by a dueling Q-network~\cite{wang2016dueling}, which separates state advantage $V_s(\cdot)$ and action advantage $A(\cdot)$, to predict the Q-values.
\begin{align}
Q_{s,a}^i=V_s(e^{[2]t}_i)+\left(A(e^{[2]t}_i)_a-\frac{1}{|\mathcal{A}|}\sum_{a'}A(e^{[2]t}_i)_{a'}\right),
\end{align}
where $\mathcal{A}$ is the action space.

After we get the $Q$ values of agent $i$, the loss function for training the whole model is a time difference (TD) error. To stabilize training, we use a multi-step TD error calculated by the mean square error (MSE)
\begin{align}
\mathcal{L}(\theta)=\mathrm{MSE}\left(R^i_t - Q^i_{s_t,a_t}(\theta)\right)
\end{align}
with $R^i_t=r^i_t+\gamma r^i_{t+1}+...+\gamma^{n}Q^i_{s_{t+n},a_{t+n}}(\bar{\theta})$, where $r^i_t$ is the reward received by agent $i$ at time $t$, $R^i_t$ is the multi-step expected return, $\theta$ represents the parameters of the entail model, and $\bar{\theta}$ denotes the parameters of the target network, a periodical copy of the online parameters $\theta$.

\subsection{Training}\label{subsec:training}
As mentioned above, we train the model from a single agent's perspective using independent Q-learning, where each agent learns its own action value $Q^i$ independently and simultaneously, treating other agents as part of the environment. The benefit is that it avoids the scalability problem in centralized training, which requires learning a Q-function for joint actions over all agents, as the joint action space grows exponentially when the number of agents increases. To further simplify the training process, instead of training separate models for different agents, we train a single model by sharing parameters among agents. To facilitate training, we adopt two training strategies, namely distributed training and curriculum learning~\cite{bengio2009curriculum}, which are discussed in the following.

\paragraph{Distributed training} Distributed training has significantly improved the performance of RL compared with non distributed ones, such as R2D2~\cite{kapturowski2018recurrent} and Ape-X~\cite{horgan2018distributed}. In MAPF, PRIMAL is a distributed RL method by using A3C. The idea behind this distributed version of algorithm is to parallelize the gradient computation, to more rapidly optimize the parameters. Another approach, as proposed in Ape-X, is to parallelize experience data generation and selection with a shared replay memory. We adopt the latter strategy, where we setup multiple runners on CPUs with their own copy of the environment and up to date model to generate experience data, and a single learner on GPU to train. The runners initialize the priorities for the experience data and feed into a global prioritized reply buffer. The learner samples the most useful experiences from the buffer to update the model parameters and also update the priorities of the experiences. The advantage is that the model training is allocated to GPU while original A3C is designed for multi-core CPU. Note that the transitions of all the agents need to be recorded for communication purpose in the model. As the priorities are shared among all runners, the good experiences explored by any runner can benefit the learner.

As we only have a single learner on GPU, the core task is to efficiently distribute multiple runners on CPUs. Here we utilize a powerful Python package named Ray~\cite{moritz2018ray}, a distributed framework designed for machine learning, to easily deploy parallel programs on multiple CPUs with little modification to the deep learning code.

\paragraph{Curriculum learning} The ultimate goal of the learning model is to handle complex cases in a large environment with high obstacle density and many agents. However, it would be hard for the model to learn fast and stably if directly starting from a complicated training environment. As pointed in~\cite{bengio2009curriculum}, presenting training examples not randomly but in a meaningful order will benefit machine learning algorithms. Motivated by this idea, we design a training pipeline to gradually introduce more complex training cases to the model. Specifically, training starts with a task where only $1$ agent stays in a $10\times 10$ environment. Then if the success rate of the current task exceeds $0.9$, two new tasks are established for training. One is to increase the number of agents by $1$, and the other is to increase the size of the environment by $5$. The final task with $16$ agents in a $40\times 40$ environment will be reached as the training scale grows.

\paragraph{Other training settings} For the environment setting, the obstacle density for each environment during training is sampled from a triangular distribution between $0$ and $0.5$ with a peak at $0.33$, which is the same as the obstacle setting in PRIMAL. The FOV size is $9\times 9$ ($10\times 10$ in PRIMAL, we make it odd) and the agent is in the middle of the FOV. The maximum step for the environment is $256$. 

For training setting, we train the model with a batch size of $128$ and a sequence length of $20$. The discount factor $\gamma=0.99$. The multi-step TD error defined in Equation 5 is computed with a length of $2$. The number of runners for distributed training is set to be $16$.

\begin{figure}[tp]
\begin{minipage}{.5\columnwidth}
\centering\
\includegraphics[scale=.256]{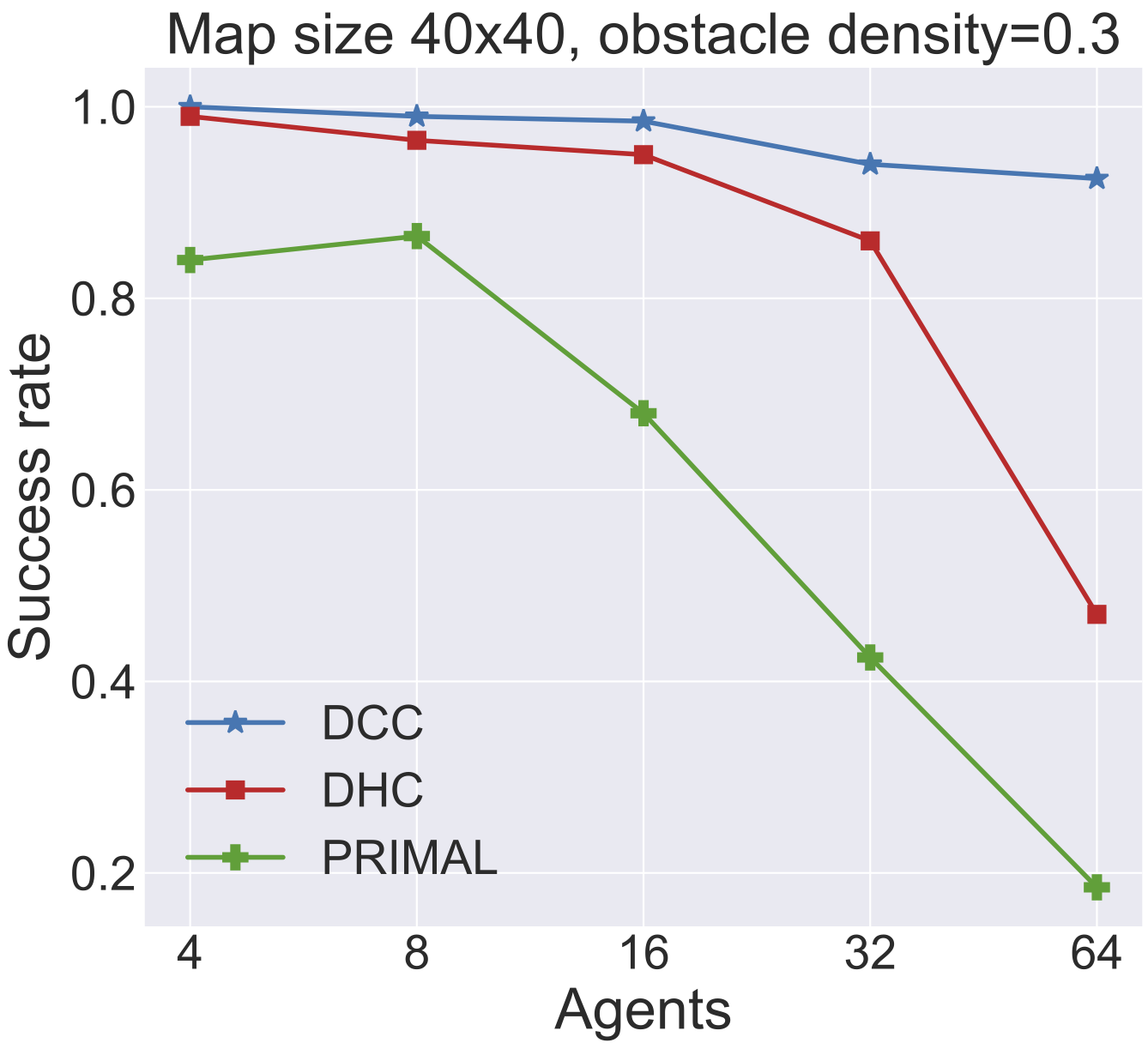}
\end{minipage}%
\begin{minipage}{.5\columnwidth}
\centering
\includegraphics[scale=.256]{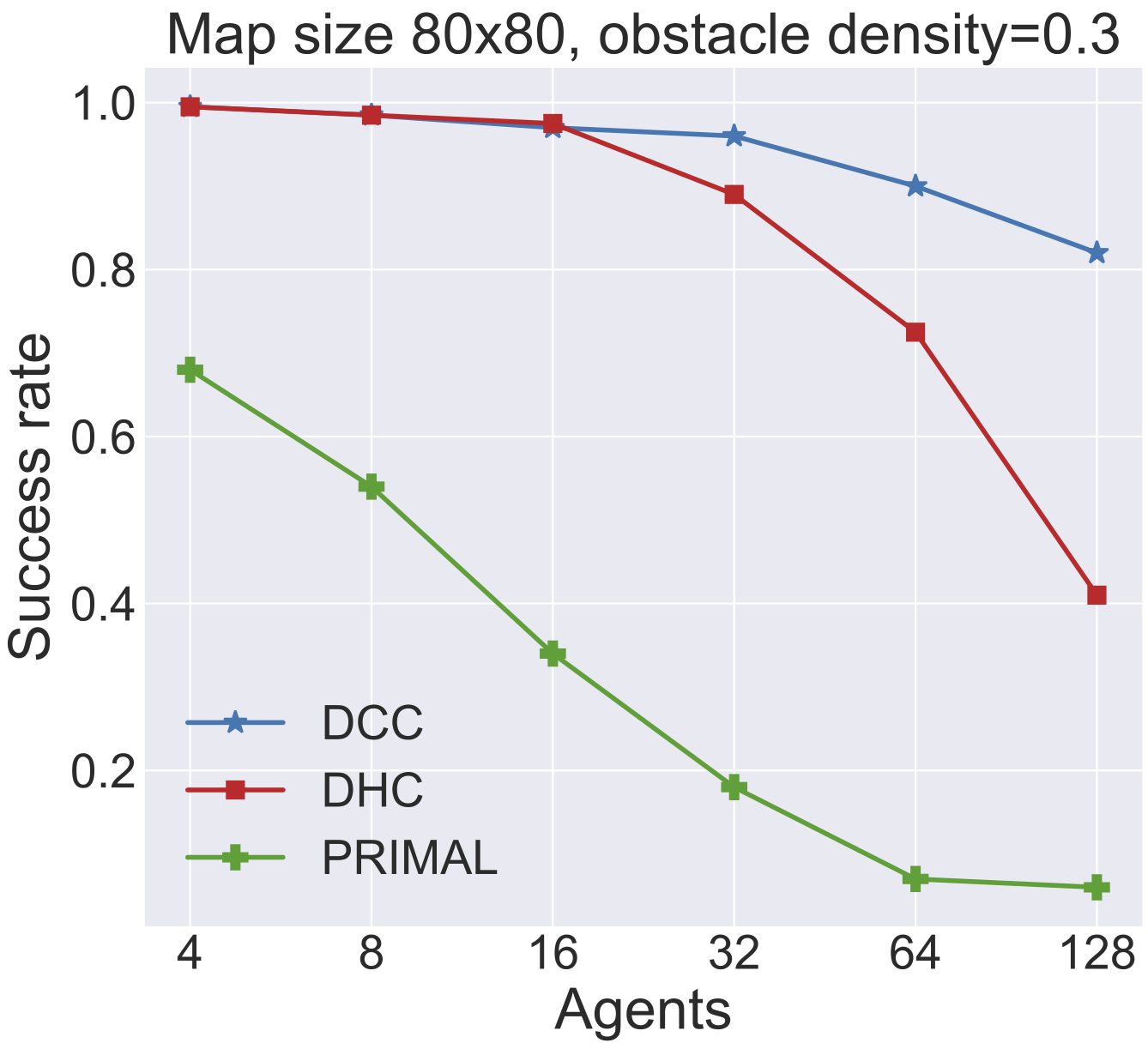}
\end{minipage}
\caption{Success rate of our method compared with PRIMAL and DHC in two different scenarios.
}
\label{fig:res_compare}
\vspace{-0.17in}
\end{figure}

\begin{table}[tbp]
\caption{Average steps in two types of environments with obstacle density $=0.3$}
\label{table:avg_step}
\begin{center}
\resizebox{0.95\columnwidth}{!}{
\begin{tabular}{c|c|ccc}

\hline
 \multicolumn{5}{c}{Average steps in $40\times 40$ maps}   \\
\hline
 Agents &  ODrM* & DCC    & DHC  & PRIMAL     \\
\hline
    4   & 50.00  & \textbf{48.575} &52.33 &79.08  \\

    8   & 52.17  & \textbf{59.60}  &63.90 & 76.53    \\

   16   & 59.78  & \textbf{71.34}  &79.63 & 107.14   \\

   32   & 67.39  & \textbf{93.54}  &100.10 & 155.21     \\

   64   & 82.60  & \textbf{135.55} &147.26 & 170.48     \\
\hline
   \multicolumn{5}{c}{}        \\
\hline
 \multicolumn{5}{c}{Average steps in $80\times 80$ maps}\\
\hline
Agents &  ODrM* & DCC    & DHC  & PRIMAL \\
\hline
    4   & 93.40  & \textbf{93.89}  &96.72 &  134.86\\

    8   & 104.92 & 109.89 &\textbf{109.24} & 153.20\\

   16   & 114.75 & \textbf{122.24} &122.54 & 180.74\\

   32   & 121.31 & \textbf{132.99}   &138.32 & 250.07\\

   64   & 134.42 & \textbf{159.67}   &163.50 & 321.63\\
   
   128 &  143.84 & \textbf{192.90}  & 213.15 &  350.76  \\
\hline

\end{tabular}
}
\end{center}
\vspace{-0.17in}
\end{table}

\section{Experiments}

The model is trained and tested in classical MAPF environments as discussed in Section~\ref{sec:env}. We compare DCC with one of the most well known RL baselines named PRIMAL~\cite{sartoretti2019primal}, and the most recent RL plus communication method named  DHC~\cite{ma2021distributed} in terms of success rate and average step. For comparing communication overhead, we build a baseline model, which uses request-reply mechanism but with predefined communication scope. PRIMAL and DHC are not included in this evaluation because they are all-to-all communication (more costly than request-reply).

\subsection{Success Rate and Average Step}
Success rate measures the ability to complete a MAPF task within given time steps. Average step measures the average time step consumed to finish a task, where smaller value indicates a better policy. We average both successful and unsuccessful cases to calculate the average steps. We set up two types of maps, $40\times 40$ and $80\times 80$ for testing. The obstacle density is set to be $0.3$, the highest testing density in PRIMAL and DHC. We set up 200 test cases for each agent number in $\{4,8,16,32,64\}$. Additional agent number of $128$ is tested in $80\times 80$ maps due to a larger environment space. The maximum time step for $40\times40$ map is 256, and 386 for $80\times80$ map, the same as the PRIMAL’s setting.

Fig.~\ref{fig:res_compare} shows the success rate of our method compared with PRIMAL and DHC in two types of environments. In general, DCC and DHC (RL plus communication) perform much better than PRIMAL (RL with expert guidance) in all cases. The performance of PRIMAL downgrades heavily in relatively larger environments ($80\times 80$) compared with smaller ones ($40\times 40$), which indicates the IL part of PRIMAL does not deliver good guidance to its RL component. Thus it suffers performance degradation on long-horizon task. In the environments where the agent density is low, such as $4$, $8$, $16$ agents in both $40\times 40$ and $80\times 80$ maps, the difference between the success rate of DCC and DHC is small. In these environments, agents have a lower chance to meet and communicate with others, so the difference between broadcast communication (DHC) and selective communication (DCC) is not very significant. However, the success rate of DHC drops rapidly when agent density further grows. In agent dense environments, agents have a higher chance of encountering each other, but agents will obtain irrelevant and redundant information from others due to broadcast communication, which harms the cooperation among agents. By selecting and gathering relevant and influential information for communication, the cooperation is enhanced by DCC.

Table~\ref{table:avg_step} verifies that DCC learns higher quality policies with respect to average steps. We include ODrM*~\cite{wagner2015subdimensional}, a centralized planner, as a reference. Especially compared with DHC, learning selective communication by DCC leads to much shorter paths in agent dense environments.

\subsection{Selective Communication vs. Predefined Scope}
Generally, selective communication by DCC does not need to set a communication scope. Any neighboring agent inside the FOV is possible to be requested by the central agent for communication. Previous broadcast communication method DHC defines the communication scope as the nearest two neighboring agents inside the FOV in order to reduce temporal redundant information. Although it is reasonable to assume the nearest agents are most relevant and influential, we show that selective communication outperforms the predefined scope (nearest two agents) for MAPF in both success rate and communication overhead.

To make a fair comparison, we develop a baseline model, named RR-N2 (request-reply, nearest two). The whole model architecture is similar to DCC. The only difference is that RR-N2 does not have the decision causal unit in Fig.~\ref{fig:model}, and in communication block, each agent only requests information from its two nearest agents inside the FOV. The training settings are the same as DCC as discussed in Section~\ref{subsec:training}.

\begin{figure}[tp]
\begin{minipage}{.5\columnwidth}
\centering\
\includegraphics[scale=.256]{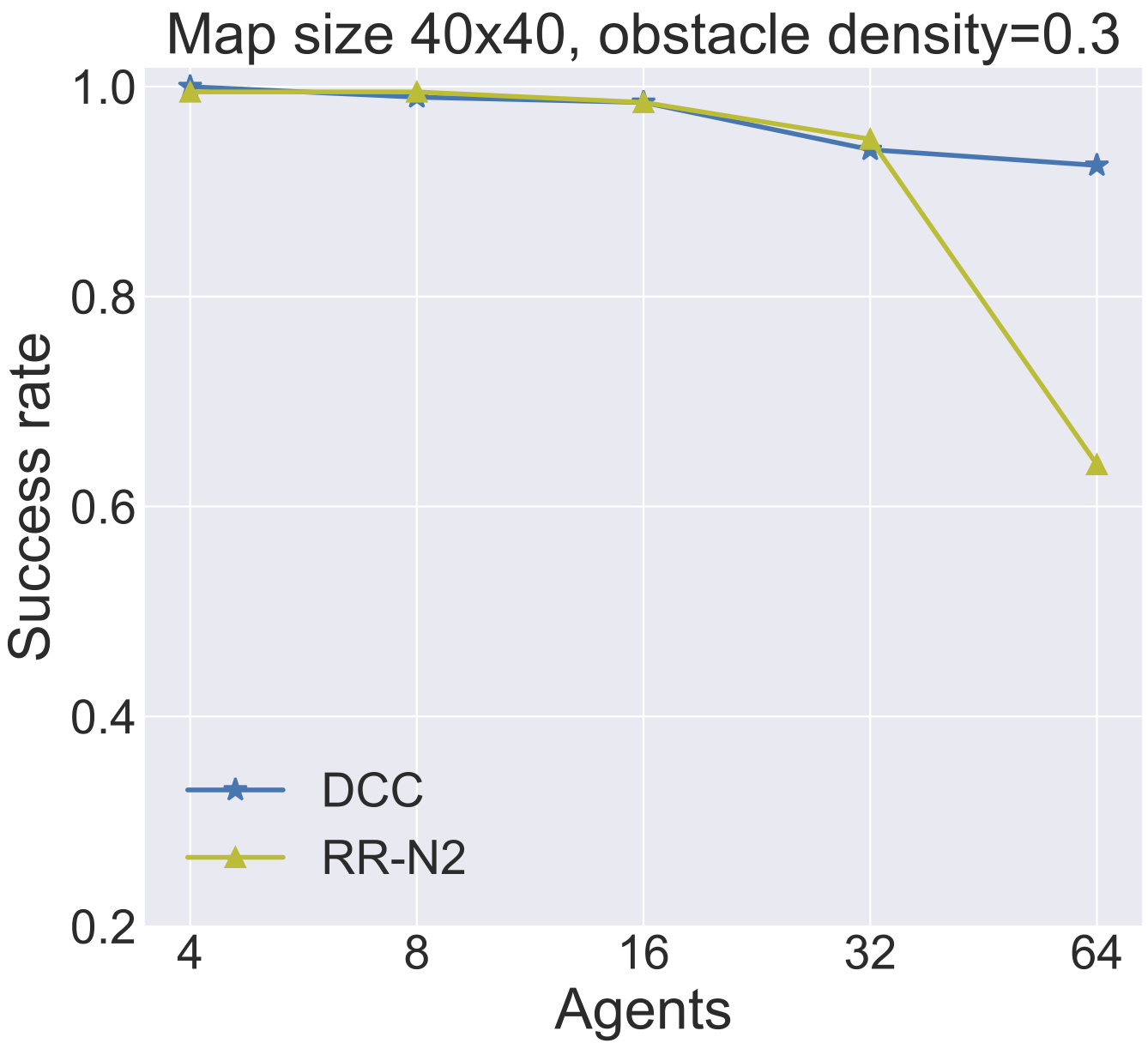}
\end{minipage}%
\begin{minipage}{.5\columnwidth}
\centering
\includegraphics[scale=.256]{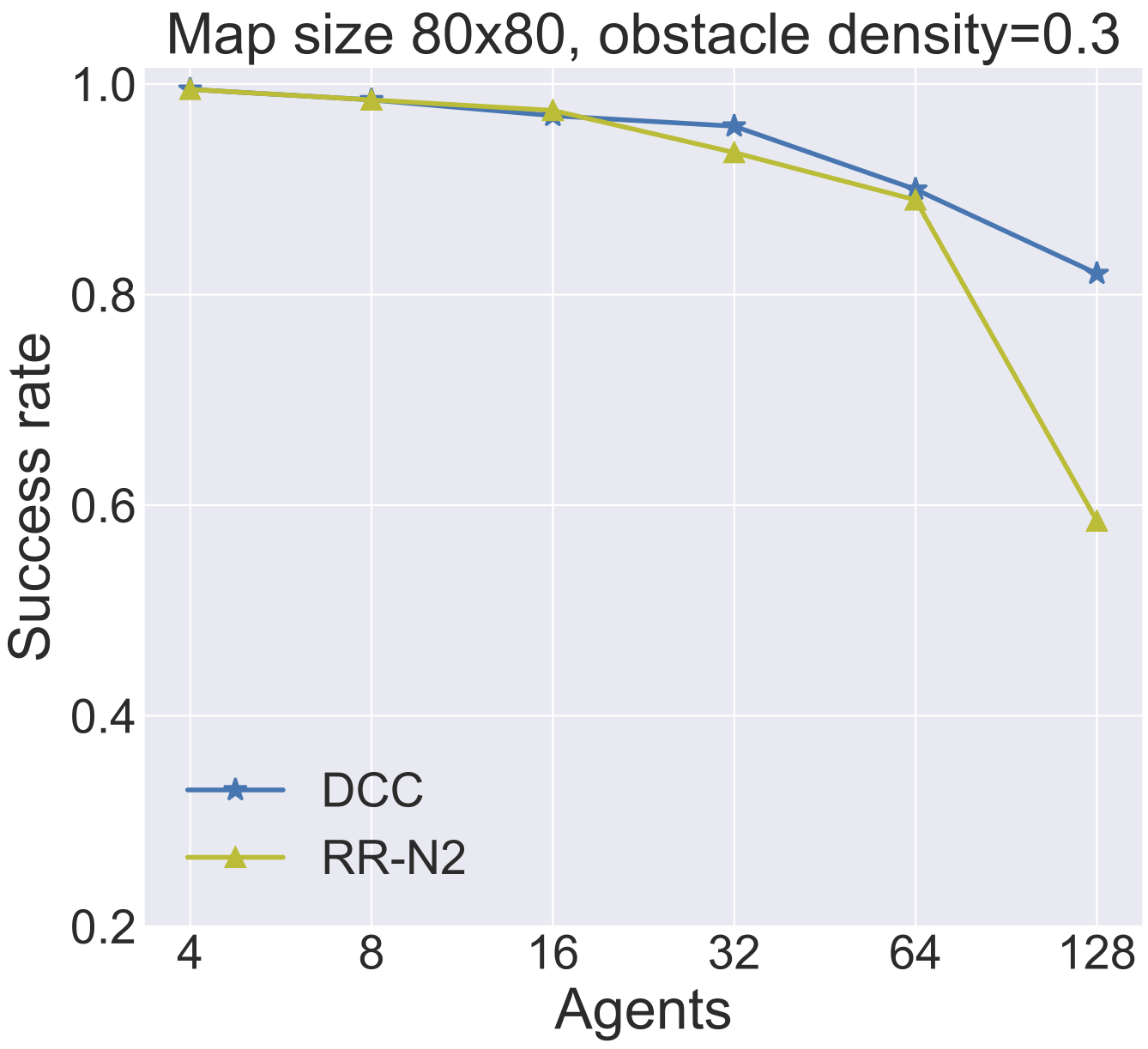}
\end{minipage}
\caption{Success rate of our method compared with RR-N2 in two different scenarios.
}
\label{fig:res_self}
\vspace{-0.17in}
\end{figure}


\begin{table}[tbp]
\caption{Communication frequency in two different environments with obstacle density $=0.3$}
\label{table:comm_freq}
\begin{center}
\resizebox{0.97\columnwidth}{!}{
\begin{tabular}{c|cc|cc}

\hline
Average Step &\multicolumn{2}{|c}{Map size $40\times 40$}      & \multicolumn{2}{|c}{Map size $80\times 80$}  \\
\hline
 Agents  & DCC     & RR-N2     & DCC     & RR-N2\\
\hline
    4    & 2.42    & 36.88     & 1.06    & 18.36 \\

    8    & 11.56   & 209.79    & 5.75    & 105.86\\
 
   16    & 60.47   & 959.38    & 24.98   & 469.58\\
 
   32    & 294.69  & 4111.57   & 126.685 & 2125.94\\

   64    & 1811.33 & 19490.09  & 562.11  & 8780.72\\
   
   128   &   -     &      -    & 2915.84 & 36560.30  \\
\hline
\end{tabular}
}
\end{center}
\vspace{-0.17in}
\end{table}

Fig.~\ref{fig:res_self} shows the success rate of our method compared with this baseline. The difference in success rate is significant in agent-dense environments, i.e., $64$ agents in $40\times 40$ map and $128$ agents in $80\times 80$ map, which implies the nearest two agents are not always the most relevant ones. One may think that DCC achieves a better performance due to unconstrained communication scope, where each agent may communicate with more than two agents inside its FOV and thus gathers more information than RR-N2. We show that this is not really the case and DCC actually learns succinct communication. In particular, we compute the average communication frequency in these test environments as shown in Table~\ref{table:comm_freq}. A pair of request and reply is counted as one time communication. The average communication overhead of RR-N2 is much more costly than DCC in all the testing cases, and is extremely expensive in agent-dense environments. By learning selective communication, DCC can greatly reduce communication by only focusing on relevant messages. 

Combining these two results, we can conclude that although communication can help multi-agent cooperation for MAPF, agents should learn to actively select relevant agents for communication, because even the most nearest agents are not the relevant ones. Communicating with only two agents can already lead to huge communication overhead. Thus, researchers should always devise a succinct communication mechanism in order to reduce the communication overhead for easy deployment to real world problems. 

{\bf Limitations:} There is an additional computational cost associated with selecting neighboring agents. In extreme cases, the cost is of order of $O(n\times p)$. $p$ is the FOV capacity.

\section{Conclusion}
We propose DCC to learn succinct communication for MAPF with classical MAPF environment settings. Our aim is to enable agents to actively select relevant and influential neighbors for communication, instead of broadcasting. The selection is learned via the local policy adjustment effect between agents, which captures the necessity of communication. Empirical results show that selective communication with relevant agents improves the policy learning process. Moreover, DCC also serves as a component for communication reduction, greatly downscaling the communication overhead. Future work entails extensions of DCC to centralized training frameworks and cooperative Markov Game tasks. Another direction is to explore how to directly exclude sets of neighbors during neighboring selection to reduce complexity, and also the maximal number of agents that DCC can handle in different size of maps.








\bibliographystyle{./IEEEtran} 
\bibliography{./IEEEabrv,./IEEEexample}

\begin{thebibliography}{10}
\providecommand{\url}[1]{#1}
\csname url@rmstyle\endcsname
\providecommand{\newblock}{\relax}
\providecommand{\bibinfo}[2]{#2}
\providecommand\BIBentrySTDinterwordspacing{\spaceskip=0pt\relax}
\providecommand\BIBentryALTinterwordstretchfactor{4}
\providecommand\BIBentryALTinterwordspacing{\spaceskip=\fontdimen2\font plus
\BIBentryALTinterwordstretchfactor\fontdimen3\font minus
  \fontdimen4\font\relax}
\providecommand\BIBforeignlanguage[2]{{%
\expandafter\ifx\csname l@#1\endcsname\relax
\typeout{** WARNING: IEEEtran.bst: No hyphenation pattern has been}%
\typeout{** loaded for the language `#1'. Using the pattern for}%
\typeout{** the default language instead.}%
\else
\language=\csname l@#1\endcsname
\fi
#2}}

\bibitem{yu2013structure}
J.~Yu and S.~M. LaValle, ``Structure and intractability of optimal multi-robot
  path planning on graphs,'' in \emph{Proc. AAAI Conference on Artificial
  Intelligence ({AAAI}'13)}, 2013.

\bibitem{banfi2017intractability}
J.~Banfi, N.~Basilico, and F.~Amigoni, ``Intractability of time-optimal
  multirobot path planning on 2d grid graphs with holes,'' \emph{IEEE Robotics
  and Automation Letters}, vol.~2, no.~4, pp. 1941--1947, 2017.

\bibitem{yu2013planning}
J.~Yu and S.~M. LaValle, ``Planning optimal paths for multiple robots on
  graphs,'' in \emph{Proc. International Conference on Robotics and Automation
  ({ICRA}'13)}.\hskip 1em plus 0.5em minus 0.4em\relax IEEE, 2013, pp.
  3612--3617.

\bibitem{surynek2016efficient}
P.~Surynek, A.~Felner, R.~Stern, and E.~Boyarski, ``Efficient sat approach to
  multi-agent path finding under the sum of costs objective,'' in \emph{Proc.
  European Conference on Artificial Intelligence ({ECAI}'16)}, 2016, pp.
  810--818.

\bibitem{goldenberg2014enhanced}
M.~Goldenberg, A.~Felner, R.~Stern, G.~Sharon, N.~Sturtevant, R.~C. Holte, and
  J.~Schaeffer, ``Enhanced partial expansion a*,'' \emph{Journal of Artificial
  Intelligence Research}, vol.~50, pp. 141--187, 2014.

\bibitem{wagner2015subdimensional}
G.~Wagner and H.~Choset, ``Subdimensional expansion for multirobot path
  planning,'' \emph{Artificial Intelligence}, vol. 219, pp. 1--24, 2015.

\bibitem{sharon2015conflict}
G.~Sharon, R.~Stern, A.~Felner, and N.~R. Sturtevant, ``Conflict-based search
  for optimal multi-agent pathfinding,'' \emph{Artificial Intelligence}, vol.
  219, pp. 40--66, 2015.

\bibitem{li2019improved}
J.~Li, A.~Felner, E.~Boyarski, H.~Ma, and S.~Koenig, ``Improved heuristics for
  multi-agent path finding with conflict-based search.'' in \emph{Proc.
  International Joint Conference on Artificial Intelligence ({IJCAI}'19)}, vol.
  2019, 2019, pp. 442--449.

\bibitem{sartoretti2019primal}
G.~Sartoretti, J.~Kerr, Y.~Shi, G.~Wagner, T.~S. Kumar, S.~Koenig, and
  H.~Choset, ``Primal: Pathfinding via reinforcement and imitation multi-agent
  learning,'' \emph{IEEE Robotics and Automation Letters (RA-L)}, vol.~4,
  no.~3, pp. 2378--2385, 2019.

\bibitem{wang2020mobile}
B.~Wang, Z.~Liu, Q.~Li, and A.~Prorok, ``Mobile robot path planning in dynamic
  environments through globally guided reinforcement learning,'' \emph{IEEE
  Robotics and Automation Letters (RA-L)}, vol.~5, no.~4, pp. 6932--6939, 2020.

\bibitem{riviere2020glas}
B.~Riviere, W.~H{\"o}nig, Y.~Yue, and S.-J. Chung, ``Glas: Global-to-local safe
  autonomy synthesis for multi-robot motion planning with end-to-end
  learning,'' \emph{IEEE Robotics and Automation Letters (RA-L)}, vol.~5,
  no.~3, pp. 4249--4256, 2020.

\bibitem{liu2020mapper}
Z.~Liu, B.~Chen, H.~Zhou, G.~Koushik, M.~Hebert, and D.~Zhao, ``Mapper:
  Multi-agent path planning with evolutionary reinforcement learning in mixed
  dynamic environments,'' in \emph{Proc. {IEEE/RSJ} International Conference on
  Intelligent Robots and Systems ({IROS}'20)}.\hskip 1em plus 0.5em minus
  0.4em\relax IEEE, 2020, pp. 11\,748--11\,754.

\bibitem{li2020graph}
Q.~Li, F.~Gama, A.~Ribeiro, and A.~Prorok, ``Graph neural networks for
  decentralized multi-robot path planning,'' in \emph{Proc. {IEEE/RSJ}
  International Conference on Intelligent Robots and Systems
  ({IROS}'20)}.\hskip 1em plus 0.5em minus 0.4em\relax IEEE, 2020, pp.
  11\,785--11\,792.

\bibitem{li2021message}
Q.~Li, W.~Lin, Z.~Liu, and A.~Prorok, ``Message-aware graph attention networks
  for large-scale multi-robot path planning,'' \emph{IEEE Robotics and
  Automation Letters (RA-L)}, vol.~6, no.~3, pp. 5533--5540, 2021.

\bibitem{ma2021distributed}
Z.~Ma, Y.~Luo, and H.~Ma, ``Distributed heuristic multi-agent path finding with
  communication,'' in \emph{Proc. {IEEE} International Conference on Robotics
  and Automation ({ICRA}'21)}, 2021.

\bibitem{littman1994markov}
M.~L. Littman, ``Markov games as a framework for multi-agent reinforcement
  learning,'' in \emph{Proc. International Conference on Machine Learning
  ({ICML}'94)}, 1994, pp. 157--163.

\bibitem{singh2018learning}
A.~Singh, T.~Jain, and S.~Sukhbaatar, ``Learning when to communicate at scale
  in multiagent cooperative and competitive tasks,'' in \emph{Proc.
  International Conference on Learning Representations ({ICLR}'19)}, 2019.

\bibitem{kim2019learning}
D.~Kim, S.~Moon, D.~Hostallero, W.~J. Kang, T.~Lee, K.~Son, and Y.~Yi,
  ``Learning to schedule communication in multi-agent reinforcement learning,''
  in \emph{Proc. International Conference on Learning Representations
  ({ICLR}'19)}, 2019.

\bibitem{zhang2019efficient}
S.~Q. Zhang, Q.~Zhang, and J.~Lin, ``Efficient communication in multi-agent
  reinforcement learning via variance based control,'' in \emph{Proc. Advances
  in Neural Information Processing Systems ({NeurIPS}'19)}, 2019, pp.
  3230--3239.

\bibitem{zhang2020succinct}
{S. Q. Zhang, Q. Zhang, and J. Lin}, ``Succinct and robust multi-agent
  communication with temporal message control,'' in \emph{Proc. Advances in
  Neural Information Processing Systems ({NeurIPS}'20)}, 2020.

\bibitem{wang2020learning}
T.~Wang, J.~Wang, C.~Zheng, and C.~Zhang, ``Learning nearly decomposable value
  functions via communication minimization,'' in \emph{Proc. International
  Conference on Learning Representations, ({ICLR}'20)}, 2020.

\bibitem{ding2020learning}
Z.~Ding, T.~Huang, and Z.~Lu, ``Learning individually inferred communication
  for multi-agent cooperation,'' in \emph{Proc. Advances in Neural Information
  Processing Systems ({NeurIPS}'20)}, 2020.

\bibitem{everett2018motion}
M.~Everett, Y.~F. Chen, and J.~P. How, ``Motion planning among dynamic,
  decision-making agents with deep reinforcement learning,'' in \emph{Proc.
  {IEEE/RSJ} International Conference on Intelligent Robots and Systems
  ({IROS}'18)}.\hskip 1em plus 0.5em minus 0.4em\relax IEEE, 2018, pp.
  3052--3059.

\bibitem{mnih2016asynchronous}
V.~Mnih, A.~P. Badia, M.~Mirza, A.~Graves, T.~Lillicrap, T.~Harley, D.~Silver,
  and K.~Kavukcuoglu, ``Asynchronous methods for deep reinforcement learning,''
  in \emph{Proc. International Conference on Machine Learning
  ({ICML}'16)}.\hskip 1em plus 0.5em minus 0.4em\relax PMLR, 2016, pp.
  1928--1937.

\bibitem{das2019tarmac}
A.~Das, T.~Gervet, J.~Romoff, D.~Batra, D.~Parikh, M.~Rabbat, and J.~Pineau,
  ``Tarmac: Targeted multi-agent communication,'' in \emph{Proc. International
  Conference on Machine Learning ({ICML}'19)}.\hskip 1em plus 0.5em minus
  0.4em\relax PMLR, 2019, pp. 1538--1546.

\bibitem{jiang2020graph}
J.~Jiang, C.~Dun, T.~Huang, and Z.~Lu, ``Graph convolutional reinforcement
  learning,'' in \emph{Proc. International Conference on Learning
  Representations ({ICLR}'20)}, 2020.

\bibitem{niu2021multi}
Y.~Niu, R.~Paleja, and M.~Gombolay, ``Multi-agent graph-attention communication
  and teaming,'' in \emph{Proc. International Conference on Autonomous Agents
  and MultiAgent Systems ({AAMAS}'21)}, 2021.

\bibitem{vaswani2017attention}
A.~Vaswani, N.~Shazeer, N.~Parmar, J.~Uszkoreit, L.~Jones, A.~N. Gomez,
  {\L}.~Kaiser, and I.~Polosukhin, ``Attention is all you need,'' in
  \emph{Proc. Advances in neural information processing systems
  ({NeurIPS}'17)}, 2017, pp. 5998--6008.

\bibitem{scarselli2008graph}
F.~Scarselli, M.~Gori, A.~C. Tsoi, M.~Hagenbuchner, and G.~Monfardini, ``The
  graph neural network model,'' \emph{IEEE transactions on neural networks},
  vol.~20, no.~1, pp. 61--80, 2008.

\bibitem{barer2014suboptimal}
M.~Barer, G.~Sharon, R.~Stern, and A.~Felner, ``Suboptimal variants of the
  conflict-based search algorithm for the multi-agent pathfinding problem,'' in
  \emph{Annual Symposium on Combinatorial Search}, 2014.

\bibitem{rashid2018qmix}
T.~Rashid, M.~Samvelyan, C.~Schroeder, G.~Farquhar, J.~Foerster, and
  S.~Whiteson, ``Qmix: Monotonic value function factorisation for deep
  multi-agent reinforcement learning,'' in \emph{Proc. International Conference
  on Machine Learning ({ICML}'18)}.\hskip 1em plus 0.5em minus 0.4em\relax
  PMLR, 2018, pp. 4295--4304.

\bibitem{stern2019multi}
R.~Stern, N.~R. Sturtevant, A.~Felner, S.~Koenig, H.~Ma, T.~T. Walker, J.~Li,
  D.~Atzmon, L.~Cohen, T.~S. Kumar, \emph{et~al.}, ``Multi-agent pathfinding:
  Definitions, variants, and benchmarks,'' in \emph{Twelfth Annual Symposium on
  Combinatorial Search}, 2019.

\bibitem{damani2021primal}
M.~Damani, Z.~Luo, E.~Wenzel, and G.~Sartoretti, ``Primal $ \_2 $: Pathfinding
  via reinforcement and imitation multi-agent learning-lifelong,'' \emph{IEEE
  Robotics and Automation Letters (RA-L)}, vol.~6, no.~2, 2021.

\bibitem{tan1993multi}
M.~Tan, ``Multi-agent reinforcement learning: Independent vs. cooperative
  agents,'' in \emph{Proc. International Conference on Machine Learning
  ({ICML}'93)}, 1993, pp. 330--337.

\bibitem{sunehag2018value}
P.~Sunehag, G.~Lever, A.~Gruslys, W.~M. Czarnecki, V.~F. Zambaldi,
  M.~Jaderberg, M.~Lanctot, N.~Sonnerat, J.~Z. Leibo, K.~Tuyls, and T.~Graepel,
  ``Value-decomposition networks for cooperative multi-agent learning based on
  team reward,'' in \emph{Proc. International Conference on Autonomous Agents
  and MultiAgent Systems ({AAMAS}'18)}, 2018, pp. 2085--2087.

\bibitem{kullback1997information}
S.~Kullback, \emph{Information theory and statistics}.\hskip 1em plus 0.5em
  minus 0.4em\relax Courier Corporation, 1997.

\bibitem{chung2014empirical}
J.~Chung, C.~Gulcehre, K.~Cho, and Y.~Bengio, ``Empirical evaluation of gated
  recurrent neural networks on sequence modeling,'' \emph{arXiv preprint
  arXiv:1412.3555}, 2014.

\bibitem{wang2016dueling}
Z.~Wang, T.~Schaul, M.~Hessel, H.~Hasselt, M.~Lanctot, and N.~Freitas,
  ``Dueling network architectures for deep reinforcement learning,'' in
  \emph{Proc. International Conference on Machine Learning ({ICML}'16)}.\hskip
  1em plus 0.5em minus 0.4em\relax PMLR, 2016, pp. 1995--2003.

\bibitem{bengio2009curriculum}
Y.~Bengio, J.~Louradour, R.~Collobert, and J.~Weston, ``Curriculum learning,''
  in \emph{Proc. International Conference on Machine Learning ({ICML}'09)},
  2009, pp. 41--48.

\bibitem{kapturowski2018recurrent}
S.~Kapturowski, G.~Ostrovski, J.~Quan, R.~Munos, and W.~Dabney, ``Recurrent
  experience replay in distributed reinforcement learning,'' in \emph{Proc.
  International Conference on Learning Representations ({ICLR}'18)}, 2018.

\bibitem{horgan2018distributed}
D.~Horgan, J.~Quan, D.~Budden, G.~Barth{-}Maron, M.~Hessel, H.~van Hasselt, and
  D.~Silver, ``Distributed prioritized experience replay,'' in \emph{Proc.
  International Conference on Learning Representations, ({ICLR}'18)}, 2018.

\bibitem{moritz2018ray}
P.~Moritz, R.~Nishihara, S.~Wang, A.~Tumanov, R.~Liaw, E.~Liang, M.~Elibol,
  Z.~Yang, W.~Paul, M.~I. Jordan, \emph{et~al.}, ``Ray: A distributed framework
  for emerging {AI} applications,'' in \emph{USENIX Symposium on Operating
  Systems Design and Implementation}, 2018, pp. 561--577.

\end{thebibliography}

\end{document}